\documentclass[letterpaper, 10 pt, conference]{ieeeconf}  %

\usepackage{balance}
\usepackage{cite}
\usepackage{makecell}

\IEEEoverridecommandlockouts                              %

\usepackage{graphicx} %
\usepackage{float} %
\usepackage{booktabs} %
\usepackage{multirow} %
\usepackage[hybrid]{markdown} %
\usepackage{amsmath} %
\usepackage{amssymb}  %
\usepackage{pifont} %
\newcommand{\xmark}{\ding{55}} %
\newcommand{\cxmark}{\multicolumn{1}{c}{\xmark}} %

\makeatletter
\let\NAT@parse\undefined
\makeatother

\usepackage{cite}
\usepackage[colorlinks,allcolors=magenta]{hyperref} %
\usepackage[all]{hypcap}

\title{\LARGE \bf
VkVIO: Cross-platform GPU Acceleration for Visual-Inertial Odometry with Vulkan
}

\author{Ole Hoffmann$^{*1}$, Mateo de Mayo$^{*1,2}$, Daniel Cremers$^{1,2}$%
\thanks{*Equal contribution.}%
\thanks{$^{1}$Ole Hoffmann, Mateo de Mayo, and Daniel Cremers are with the Technical University of Munich, Munich, Germany.\newline
        {\tt\small \{ole.hoffmann, mateo.demayo, cremers\}@tum.de}}%
\thanks{$^{2}$Mateo de Mayo and Daniel Cremers are also with the Munich Center for Machine Learning, Munich, Germany.}%
}

\newread\imgstream
\immediate\openin\imgstream=imagedata.in
\makeatletter
\def\new@kvginclip#1{}
\def\new@kvgintrim#1{}
\let\old@kvginclip\KV@Gin@clip
\let\old@kvgintrim\KV@Gin@trim
\let\oldincludegraphics\includegraphics
\providecommand{\includegraphics}{}
\renewcommand{\includegraphics}[2][]{%
  \immediate\read\imgstream to \src
  \immediate\read\imgstream to \removecrop
  \ifnum\removecrop=1
      \let\KV@Gin@clip\new@kvginclip
      \let\KV@Gin@trim\new@kvgintrim
  \fi
  \oldincludegraphics[#1]{\src}%
  \let\KV@Gin@clip\old@kvginclip
  \let\KV@Gin@trim\old@kvgintrim}
\makeatother

\begin{document}

\maketitle
\thispagestyle{empty}
\pagestyle{empty}

\begin{abstract}
Perception in robotics and XR fundamentally relies on good state estimation. Visual-inertial odometry (VIO) and Simultaneous Localization and Mapping (VI-SLAM) are proven ways of achieving this goal in a cost-effective and accurate manner. Efficiency in these systems allows for smaller, cooler, and lighter devices. GPU acceleration is a natural approach for reducing latency, thanks to their wide availability in platforms like embedded computers, mobile phones, and XR headsets. However, previous works in the literature have limited themselves to the use of CUDA for this task, significantly reducing deployment options to a single vendor. We instead leverage the vendor-agnostic Vulkan API, originally designed for the strict performance requirements of 3D graphics applications. In this work, we present VkVIO, the first, to the best of our knowledge, cross-platform GPU-accelerated VIO method. We provide state-of-the-art accuracy with causal estimates required for real-time operation. We deploy VkVIO on a diverse range of devices spanning a workstation, a laptop, and an extremely inexpensive single-board computer, while outperforming CUDA-based systems on the same hardware. VkVIO enables possibilities for low-latency, low-power, and low-cost VIO in robotics and XR.
\end{abstract}

\section{INTRODUCTION}

Perception capabilities are foundational to robotic systems \cite{lavalleVirtualRealityEmerging2024a} and require good pose estimation for navigating and interacting with their environment. Visual-inertial odometry (VIO) and Simultaneous Localization and Mapping (VI-SLAM) have proven to be a cost-effective and accurate method for this purpose, leveraging cameras and inertial measurement units (IMU) to provide pose estimates \cite{scaramuzzaVisualOdometryTutorial2011, cadenaPresentFutureSimultaneous2016}. It is important to optimize the performance of these estimators to reduce power consumption, heat, and latency, while improving battery life, weight, and device size. Given the widespread availability of GPUs across platforms, they are a natural choice for accelerating this task \cite{herten2023manycores}.

\begin{figure}[t]
    \centering
    \includegraphics[width=\columnwidth]{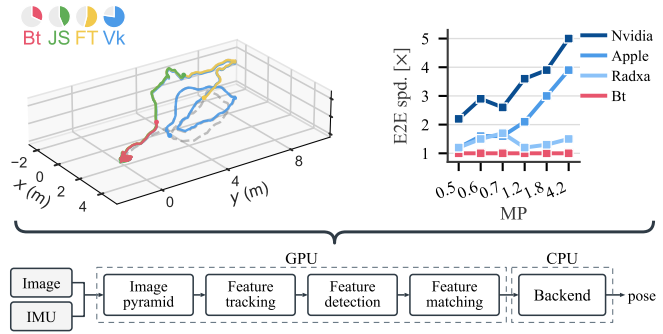}
    \caption{\textbf{VkVIO} runs the frontend of a stereo visual-inertial
    odometry (VIO) pipeline on any Vulkan-capable GPU. \emph{Top left:} progress
    on EuRoC MH02 after the same wall-clock time for VkVIO (Vk), Basalt (Bt),
    Jetson-SLAM (JS), and FastTrack (FT), all executed on the same
    Nvidia~RTX~3070 machine. \emph{Top right:} end-to-end speedup of VkVIO over Basalt on Nvidia~RTX~3070, Apple~M3, and Radxa~Cubie~A7Z for six evaluation sequences, ordered by the pixels each delivers per frame (all cameras, in megapixels). \emph{Bottom:} general pipeline overview of VkVIO.}
    \label{fig:teaser}
\end{figure}

GPUs are well suited to the parallel workloads found in VIO/VI-SLAM \cite{garland2010throughput, owens2008gpucomputing}, as evidenced in prior work \cite{nagyFasterFASTGPUAccelerated2020a, khabiriFastTrackGPUAcceleratedTracking2025a, kumarHighSpeedStereoVisual2024}, but these approaches are CUDA-based. This leaves all other GPU vendors out, meaning these techniques cannot be leveraged on a wide range of devices \cite{herten2023manycores}, including all mobile phones, most single-board computers, and laptops without CUDA GPUs. Furthermore, CUDA-capable devices tend to target higher power envelopes and price ranges than alternatives, which makes them less ideal for low-power, low-cost embedded devices \cite{kumarHighSpeedStereoVisual2024, lu2022parallel}. To exemplify the potential of cross-platform GPU acceleration, in this work we accelerate a \textit{Radxa Cubie A7Z}, a single-board computer (SBC) that is at least 20$\times$ cheaper and has lower power consumption than the closest CUDA-capable alternative in the Jetson series.

While cross-platform GPU compute APIs such as OpenCL or OpenVX exist, they are usually not as well supported, or provide worse performance than CUDA on Nvidia hardware \cite{karimiPerformanceComparisonCUDA2011}, which remains highly prevalent in scientific domains. We propose instead the use of Vulkan \cite{khronos_vulkan}, a low-level, cross-platform graphics and compute \cite{VComputeBench} API supported across most GPU vendors. Vulkan was originally designed for the performance needs of graphical applications like games, but has since been extended to general-purpose compute, with excellent support across embedded devices like Raspberry Pi and Radxa boards, mobile phones, laptops, and virtually all GPU manufacturers.

Perception algorithms are used across many domains and applications; while their constraints vary, most require tracking capabilities and have some GPU available. A cross-platform GPU implementation lets the same hardware-accelerated state-estimation codebase deploy across all of them without vendor-specific rewrites. Vulkan is also usually among the first APIs supported by new GPU platforms, compared to OpenCL or OpenVX, and remains an important performance target on Nvidia hardware due to the gaming market. This flexibility lets practitioners pick the best hardware for a task, upgrade it during development, and run the same codebase on workstations and deployed embedded devices alike.

Given this context, the contributions of this work are:

\begin{itemize}
    \item VkVIO, the first, to the best of our knowledge, cross-platform GPU-accelerated VIO frontend, built on Vulkan/SPIR-V instead of CUDA, shown running across Nvidia, Apple, and ARM-class GPUs.
    \item A reformulation of the CPU-based VIO frontend of Basalt \cite{usenkoBasaltVisualInertialMapping2020,demayoMonadoSLAMDataset2025} into Vulkan compute shaders, including subgroup-based and shared-memory fallback variants for hardware without subgroup support, released as open source\footnote{Source code will be released}.
    \item A cross-platform evaluation of three devices spanning vendors and price/power points (Nvidia RTX 3070 workstation, Apple M3 laptop, Radxa Cubie A7Z SBC), showing frontend speedups of as much as 11$\times$ over the CPU baseline, more stable per-frame latency, and lower power and thermal load, all without loss of accuracy.
    \item A comparison against CUDA-only accelerated frontends (Jetson-SLAM, FastTrack) on the same Nvidia hardware, showing that VkVIO's vendor-agnostic approach even surpasses their performance.
\end{itemize}

\section{RELATED WORK}

The pipeline of VIO and VI-SLAM systems can be separated into two components \cite{cadenaPresentFutureSimultaneous2016}: a visual frontend that converts raw images into keypoints and a backend that fuses them with inertial data to estimate the system's position. Existing approaches can be broadly categorized into filter-based, such as MSCKF \cite{mourikisMultiStateConstraintKalman2007a} and ROVIO \cite{bloeschRobustVisualInertial2015}, or optimization-based, such as OKVIS \cite{leuteneggerKeyframeBasedVisualInertialSLAM2013}, VINS-Mono \cite{qinVINSMonoRobustVersatile2018}, ORB-SLAM3 \cite{camposORBSLAM3AccurateOpenSource2021}, and Basalt \cite{usenkoBasaltVisualInertialMapping2020}. Our work builds on Basalt with modifications from \cite{demayoMonadoSLAMDataset2025}.

The visual frontend is an attractive target for GPU acceleration since it processes every incoming frame, and its workload grows with image resolution, camera count, and number of tracked features. Its operations are inherently data-parallel, applying the same instruction across many pixels or features, and thus map naturally onto the GPU's massively parallel architecture \cite{garland2010throughput, owens2008gpucomputing}.

Previous work has demonstrated that moving frontend operations to the GPU can improve performance. However, most existing methods rely on CUDA, Nvidia's proprietary API, tying them to Nvidia hardware and limiting deployment across other GPU vendors.

Data Flow ORB-SLAM \cite{aldegheriDataFlowORBSLAM2019} restructures the feature extraction pipeline of ORB-SLAM2 \cite{mur-artalORBSLAM2OpenSourceSLAM2017}, offloading image-pyramid construction and FAST corner detection to the GPU. Muzzini et al.~\cite{muzziniBriefAnnouncementOptimized2023} extend this by additionally offloading keypoint orientation and ORB descriptor computation. Other systems extend acceleration beyond feature extraction. Faster than FAST \cite{nagyFasterFASTGPUAccelerated2020a} accelerates feature detection and tracking for monocular VIO, Jetson-SLAM \cite{kumarHighSpeedStereoVisual2024} introduces a fully CUDA-accelerated frontend for stereo SLAM, and FastTrack \cite{khabiriFastTrackGPUAcceleratedTracking2025a} reuses an existing CUDA implementation of ORB extraction and additionally accelerates stereo matching and search-by-projection during local-map tracking.

GLidE-SLAM \cite{desousaGLidESLAMGLAcceleratedIndirectDirect2026a}, a recent work close to ours in spirit, uses OpenGL compute shaders, a cross-platform API, to accelerate direct photometric tracking on intermediate frames, showing that VI-SLAM can benefit from GPU acceleration without CUDA. Its implementation is, however, monocular only and thus not directly comparable to our work. While OpenGL provides portability across GPU vendors, we argue that Vulkan is the better choice, given its growing importance as a graphics API on modern platforms and continued vendor investment driven by the gaming market.

Vulkan's suitability for high-performance, vendor-independent computation is further supported by work outside VIO and VI-SLAM \cite{VComputeBench}: VkFFT \cite{tolmachevVkFFTAPerformantCrossPlatform2023} provides a cross-platform alternative to vendor-specific FFT libraries, and VkSplat \cite{chen2026vksplat} implements a 3D Gaussian Splatting training pipeline in Vulkan compute across GPU vendors. Neither addresses VIO directly, but both demonstrate the viability of Vulkan for complex, high-throughput pipelines.

We therefore adopt Vulkan as our cross-platform compute API and introduce VkVIO, a GPU-accelerated implementation of Basalt's frontend. To the best of our knowledge, this is the first work to use Vulkan for cross-platform GPU acceleration of stereo VIO.

\section{METHOD}
\begin{figure}[t]
    \centering
    \includegraphics[width=\columnwidth]{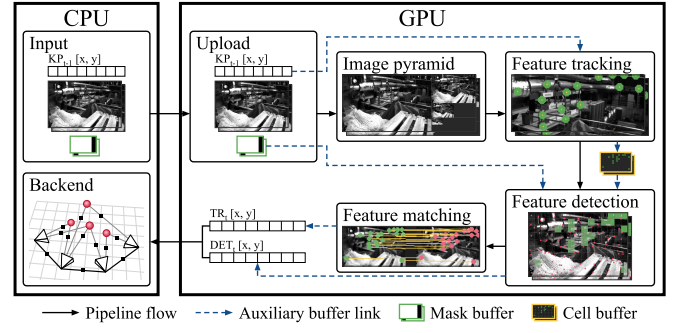}
    \caption{Data flow of the VkVIO frontend for one frame. The CPU uploads the
stereo images, the previous keypoints $\mathrm{KP}_{t-1}$, and the mask
buffer. The four stages run as compute shaders from a single command buffer.
The outputs $\mathrm{TR}_t$ (tracked and cross-matched keypoint positions)
and $\mathrm{DET}_t$ (newly detected keypoints) are downloaded back to the CPU.}
    \label{fig:frontend}
\end{figure}

\subsection{Architecture}
Vulkan is a cross-platform API for graphics and compute on modern GPUs. Its explicit programming model gives applications direct control over resource management, synchronization, and command submission, enabling high performance across many platforms \cite{khronos_vulkan}.

Compute work is issued as dispatch commands, which are recorded into a command buffer and submitted to a queue for execution. Each dispatch runs the compute pipeline bound at recording time and launches \texttt{groupCountX} $\times$ \texttt{groupCountY} $\times$ \texttt{groupCountZ} workgroups. A workgroup consists of invocations that execute the compute shader. Within a workgroup, invocations are further partitioned into subgroups, sets of invocations that the hardware issues together on a single SIMD unit. The subgroup size is a device property. Members of a subgroup can exchange values directly via subgroup operations, avoiding the shared memory and barriers that would otherwise be required.

All shaders in Vulkan are defined as SPIR-V, a hardware-independent binary intermediate representation. Shaders are written in GLSL and compiled to SPIR-V ahead of time. At pipeline creation, the driver translates the compiled module into native instructions for the target GPU.

\subsection{Pipeline}
The Basalt frontend consists of four stages: pyramid construction, temporal feature tracking, feature detection, and cross-camera matching. VkVIO implements each stage using a Vulkan compute shader and records the per-frame workload into a single command buffer, which is submitted to the compute queue once per frame. Consecutive stages are separated by pipeline barriers.
All device resources, including pyramid images, keypoint buffers, and mask and cell buffers, are allocated once during initialization and reused across frames. After submitting the command buffer, the host waits on a fence for GPU execution to complete, then reads the detected keypoints and tracking results from host-visible memory. The following subsections describe the implementation of each stage.

\subsection{Pyramid}
The first stage of the frontend builds an image pyramid from each incoming camera frame \cite{bouguet2001pyramidalf}. Each level $\ell+1$ is constructed by a separable $5 \times 5$ binomial blur followed by $2:1$ subsampling,
\begin{align}
L_{\ell+1}(x,y) = \Big\lfloor \tfrac{1}{256}\sum_{i=-2}^{2}\sum_{j=-2}^{2} w_i\, w_j\, L_\ell(2x+i,\,2y+j) \Big\rceil,
\label{eq:pyramid}
\end{align}
with $w=[1,4,6,4,1]$ the binomial kernel indexed from $-2$ to $2$ and $\lfloor\cdot\rceil$ rounding to the nearest integer. Since the kernel is separable, the downsampling splits into a horizontal and a vertical pass,
\begin{align}
V_{\ell}(x,y) &= \sum_{i=-2}^{2} w_i\, L_{\ell}(2x+i,\, y),
\label{eq:pyramid_horizontal}\\
L_{\ell+1}(x,y) &= \Bigg(\sum_{j=-2}^{2} w_j\, V_{\ell}(x,\, 2y+j)\Bigg) \gg 8,
\label{eq:pyramid_vertical}
\end{align}
with $V_{\ell}$ the intermediate result of the horizontal pass.

On the GPU, one invocation computes one pixel of the destination level $\ell+1$. Let $S$ denote the subgroup size of the target device. A workgroup consists of $S \times R$ invocations and emits an $S \times R$ output tile, so that each row of the tile maps onto exactly one subgroup and executes in lockstep. To produce its tile, a workgroup reads a $(2S+4) \times (2R+4)$ region of level $\ell$. It first evaluates the horizontal pass \eqref{eq:pyramid_horizontal} over this region, yielding $S \times (2R+4)$ intermediate values $V_\ell$. After a barrier, each invocation evaluates the vertical pass \eqref{eq:pyramid_vertical} and writes its pixel into level $\ell+1$. Since $S$ varies between vendors, the tile dimensions are configured per device. On our hardware, $S = 32$, and we choose $R = 8$, giving workgroups of $256$ invocations, $32 \times 8$ output tiles, and $68 \times 20$ source regions.

Each level is produced by one dispatch whose tiles cover the level entirely. Since each level depends on the output of the previous one, dispatches run in sequence, and parallelism is confined within each level. All levels of one camera are stored in a single two-dimensional texture atlas (Fig.~\ref{fig:frontend}), from which the tracking and matching stages read.
\subsection{Detection}
\begin{figure}[t]
    \centering
    \includegraphics[width=\columnwidth]{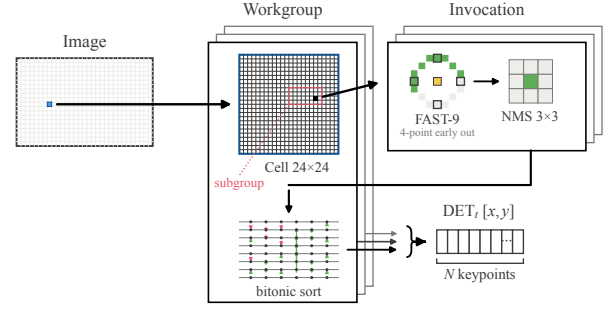}
    \caption{GPU feature detection, from image to invocation. Each $24\times24$ grid cell (blue) maps to one workgroup, its pixels to subgroups of 32 invocations (pink), and each pixel (black) to one invocation, which runs the four-point early-out, the FAST-9 segment test, and $3\times3$ NMS. A bitonic sort per workgroup keeps the $k$ strongest survivors, appended to $\mathrm{DET}_t$.}
    \label{fig:detector}
\end{figure}
Following Basalt, we partition the image into cells with a per-cell feature limit, a grid-based selection strategy commonly used in VIO and VI-SLAM to encourage a spatially uniform keypoint distribution. Within each cell, candidates are evaluated with the FAST-9 corner detector \cite{rostenMachineLearningHighSpeed2006} which defines a pixel $p$ as a corner if at least nine contiguous pixels on a sixteen-pixel Bresenham circle of radius three are all brighter than $I(p)+\tau$ or all darker than $I(p)-\tau$, and the corner score is the largest $\tau$ for which this holds.

Each cell is processed independently over the threshold sequence $\langle 40,20,10,5\rangle$. At each threshold, corner scores are computed, non-maximum suppression (NMS) is applied, and the strongest $k$ survivors are selected by sorting. If fewer than $k$ keypoints are found, the procedure repeats at the next threshold until the target count is reached or the sequence is exhausted. In our evaluation, we use $k=1$, the default setting of Basalt.

On the GPU, each grid cell is assigned to one workgroup whose size is configured independently for the target hardware. Depending on cell dimensions and workgroup size, each invocation processes one or several pixels.

Each workgroup first loads its image tile cooperatively into shared memory to avoid repeated global-memory accesses, then evaluates the FAST response for its assigned pixels. As shown in Fig.~\ref{fig:detector}, an early-rejection test on the opposite ring-point pairs $(0,8)$ and $(4,12)$ discards candidates whose comparisons exclude both bright and dark FAST-9 arcs before maximum-threshold scoring. A $3\times3$ NMS stage then retains only strict local maxima, and a bitonic $k$-sort network selects up to $k$ keypoints with the highest scores per cell.

For embedded GPUs with limited execution resources, such as the Radxa platform used in our evaluation, we provide a variant specialized for single-keypoint selection. It replaces the bitonic $k$-sort with an atomic maximum reduction over the NMS survivors, yielding the highest-scoring candidate per cell without the repeated compare-exchange operations and workgroup barriers of the sorting network. Both variants write the selected keypoint coordinates and scores to an output buffer that is accessible by the CPU.

\subsection{Tracking}
\begin{figure}[t]
    \centering
    \includegraphics[width=\columnwidth]{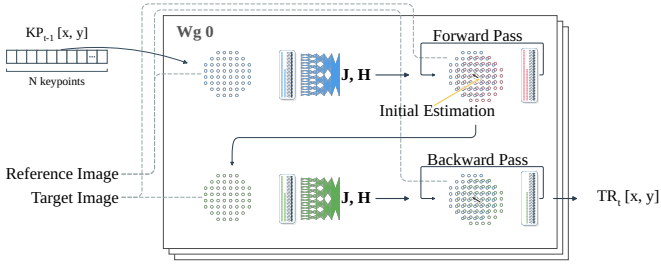}
    \caption{GPU feature tracking, one workgroup per keypoint. \emph{Forward pass:} the reference patch (blue) is sampled from the reference image. The backend's initial estimate (pink) places it in the target image, where it is iterated until it aligns with the reference patch (blue). \emph{Backward pass:} the same steps with the images exchanged (green). The returned position is validated against the original and written to the output buffer $\mathrm{TR}_t$. Dashed connectors indicate which image each block reads.}
    \label{fig:tracking}
\end{figure}
For keypoint tracking over consecutive frames, as well as for matching keypoints across cameras, we use a patch-based, inverse-compositional KLT approach~\cite{lucasIterativeImageRegistration1981,Baker-2002-8493}. The displacement of a feature between consecutive frames, as well as between cameras, is modeled as a planar rigid transform $\mathbf{T}(\boldsymbol{\xi})\in SE(2)$ with $\boldsymbol{\xi}=(t_x,t_y,\theta)^{\mathsf T}\in\mathbb{R}^{3}$, which is recovered by minimizing the photometric error over a fixed set of $52$ sample locations $\Omega=\{\mathbf{x}_i\}$ arranged around the feature. To remove any global brightness offset between frames, each patch is divided by its own mean intensity $\bar{I}$, giving us the residual
\begin{equation}
e_i(\boldsymbol{\xi})
  = \frac{I_{t+1}\big(\mathbf{T}(\boldsymbol{\xi})\,\mathbf{x}_i\big)}{\bar{I}_{t+1}}
  - \frac{I_{t}(\mathbf{x}_i)}{\bar{I}_{t}},
  \qquad \mathbf{x}_i\in\Omega ,
\label{eq:klt_residual}
\end{equation}
which is driven toward zero by repeated Gauss-Newton steps
\begin{equation}
\boldsymbol{\delta} = -\mathbf{G}\,\mathbf{e},
\qquad
\mathbf{T} \leftarrow \mathbf{T}\circ\exp(\boldsymbol{\delta}) .
\label{eq:klt_update}
\end{equation}
Here, $\mathbf{e}$ is the residual vector, and $\mathbf{J}$ is its Jacobian with respect to the increment $\boldsymbol{\delta}$. In the inverse-compositional formulation, $\mathbf{J}$ is evaluated on the source image, so it does not change between iterations. Its $i$-th row is $\mathbf{J}_i=[\,g_x,\;g_y,\;g_y u_i - g_x v_i\,]$, where $(g_x,g_y)$ is the gradient of the source image $\tilde{I}_t$ at the sample $\mathbf{p}+\mathbf{x}_i$, with $\mathbf{p}$ being the keypoint and $\mathbf{x}_i$ being the pattern offset. From $\mathbf{J}$, we form the Gauss-Newton Hessian $\mathbf{H}=\mathbf{J}^{\mathsf T}\mathbf{J}=\sum_i \mathbf{J}_i^{\mathsf T}\mathbf{J}_i$ and the gain $\mathbf{G}=\mathbf{H}^{-1}\mathbf{J}^{\mathsf T}$, both of which are computed only once.

Each tracking call consists of a forward and a backward pass, shown in Fig.~\ref{fig:tracking}. The forward pass first builds the template from the source image, gathering the samples and their gradients once per feature and pyramid level. The warp is then initialized by projecting the pattern into the target image through the camera model, using the predicted pose and inverse depth, and refined by descending the pyramid coarse to fine with five Gauss-Newton iterations per level. The backward pass repeats this with the roles of the two images exchanged, rebuilding the template on the target image and initializing the warp from the forward estimate. Finally, the feature is discarded if the distance between the returned position and the original one exceeds a fixed threshold.

We dispatch one workgroup per keypoint, so that the invocations of a workgroup collectively process a single keypoint with the work divided across the patch since the Hessian $\mathbf{H}$ and the increment $\boldsymbol{\delta}$ in \eqref{eq:klt_update} are sums over the $52$ points of $\Omega$, and each invocation accumulates a share of those points. We size the workgroup to exactly one subgroup of $32$ invocations, so that the partial sums can be combined with subgroup reductions. Subgroup arithmetic is, however, an optional capability that not every GPU exposes, particularly low-end devices such as the Radxa Cubie A7Z used in our evaluation. For such platforms, we provide a second shader variant that replaces the subgroup reductions with an accumulation through shared memory, at the cost of additional barriers.

\section{RESULTS}
\begin{figure}[t]
    \centering
    \includegraphics[width=\columnwidth]{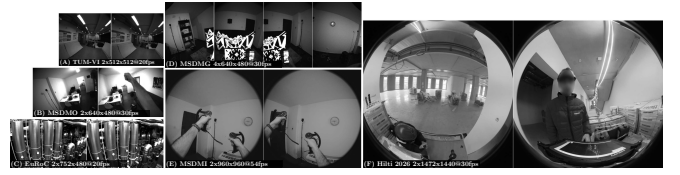}
    \caption{Input imagery of the six capture devices in the evaluation.
    Each panel shows one frame from every camera of a device, labeled with
    camera count, resolution, and frame rate, and all panels are drawn to a
    common pixel scale so that their area reflects the pixels delivered per
    frame.}
    \label{fig:datasets}
\end{figure}
\subsection{Dataset Selection}

For dataset selection, we prioritize a broad selection of devices over sequence count, to showcase the versatility of VkVIO and how its performance scales with input resolution. We utilize visual-inertial datasets from the SLAM literature centered around real-time robotics and XR. Fig.~\ref{fig:datasets} shows an overview of the type of input data provided by each device in the datasets, with the pixel area of each device drawn to scale.

The EuRoC dataset \cite{burriEuRoCMicroAerial2016a} provides images from a micro aerial vehicle flying indoors, while the TUM-VI dataset \cite{schubertBasaltTUMVI2018} provides images from a handheld rig. The Monado SLAM dataset (MSD) \cite{demayoMonadoSLAMDataset2025} provides images from three different VR headsets with different input resolutions, one of them with four cameras instead of two, all of which VkVIO is able to use. Finally, the Hilti Challenge 2026 dataset \cite{slamchallenge2026} provides 360$^\circ$ field-of-view (FoV) coverage from two wide-FoV cameras facing opposite directions.

We choose one representative sequence per capture device, since per-frame timing metrics have little variation across sequences from the same device. These are the EuRoC Machine Hall 02 sequence (EMH02), the TUM-VI Room 2 sequence (TR2), the MSD sequences MOO02, MGO02, and MIO02 (three different devices), and, for Hilti 2026, the \texttt{floor1/2025-05-05/run1} sequence (HF151). All figures that compare
input resolutions refer to these sequences.

\subsection{Improvements over CPU-Only Baseline}
\begin{table}[t]
    \centering
    \footnotesize
    \caption{Accuracy maintained from Basalt baseline}
    \label{tab:accuracy}
    \setlength{\tabcolsep}{3.5pt}
\begin{tabular}{llrrrrrr}
  \toprule
   &  & \multicolumn{2}{c}{\textbf{ATE (m)}} & \multicolumn{2}{c}{\textbf{RTE (m)}} & \multicolumn{2}{c}{\textbf{Time (ms)}} \\
  \cmidrule(lr){3-4}\cmidrule(lr){5-6}\cmidrule(lr){7-8}
  \textbf{Seq.} & \textbf{Sys.} & \textbf{mean} & \textbf{med.} & \textbf{mean} & \textbf{med.} & \textbf{FE} & \textbf{E2E} \\
  \midrule
  MOO* & Bt & 0.213 & \textbf{0.063} & 0.009 & 0.008 & $3.6\pm0.2$ & $4.8\pm0.3$ \\
  {\scriptsize 16 seq.} & Vk  & \textbf{0.160} & 0.064 & 0.009 & 0.008 & {\boldmath$1.9\pm0.1$} & {\boldmath$3.1\pm0.3$} \\
  \midrule
  MGO* & Bt & \textbf{0.220} & \textbf{0.075} & 0.014 & 0.013 & $7.0\pm0.4$ & $8.6\pm0.3$ \\
  {\scriptsize 15 seq.} & Vk  & 0.221 & 0.080 & 0.014 & 0.013 & {\boldmath$3.0\pm0.2$} & {\boldmath$4.6\pm0.5$} \\
  \midrule
  MIO* & Bt & \textbf{0.312} & \textbf{0.088} & 0.014 & 0.007 & $8.4\pm0.4$ & $9.7\pm0.5$ \\
  {\scriptsize 16 seq.} & Vk  & 0.313 & 0.092 & 0.014 & 0.007 & {\boldmath$2.2\pm0.1$} & {\boldmath$3.5\pm0.2$} \\
  \midrule
  \multirow{2}{*}{Mean} & Bt & 0.249 & \textbf{0.068} & 0.012 & 0.008 & $6.3\pm2.1$ & $7.7\pm2.2$ \\
                         & Vk  & \textbf{0.231} & 0.080 & 0.012 & 0.008 & {\boldmath$2.3\pm0.5$} & {\boldmath$3.7\pm0.7$} \\
  \bottomrule\vspace{-5pt}
\end{tabular}
\scriptsize\raggedright
\textit{\textbf{ATE/RTE}~absolute/relative trajectory error. $SE(3)$ aligned/$\Delta=6$ frames\quad\textbf{FE/E2E}~frontend/end-to-end time per frame, mean\,$\pm$\,standard deviation in ms}
\end{table}
We compare Basalt CPU frontend against VkVIO GPU frontend. Both share the unchanged Basalt backend, the same sequences, and the same machine, so the differences stem from moving the frontend to the GPU. Frontend (FE), backend (BE), and end-to-end (E2E) times are host wall-clock intervals per frame unless stated otherwise. Table~\ref{tab:accuracy} reports full-pipeline accuracy and per-frame times over 47 MSD sequences on the Apple M3. The frontend speedup grows with the pixels per frame (1.9$\times$ on MOO, 2.3$\times$ on MGO, and 3.8$\times$ on MIO), and the end-to-end speedup, diluted by the unchanged backend, follows at 1.5$\times$, 1.9$\times$, and 2.8$\times$. Furthermore, accuracy is preserved in every sequence group as shown by the median absolute trajectory error (ATE) agreeing within 6 mm, and the relative trajectory error (RTE) being identical at millimeter precision in both mean and median. The larger gap in mean ATE on MOO, favoring VkVIO, comes from a few sequences where the two runs drift apart. Since both systems run the same algorithm on every frame, we attribute it to floating-point arithmetic.
\begin{figure}[t]
    \centering
    \includegraphics[width=\columnwidth]{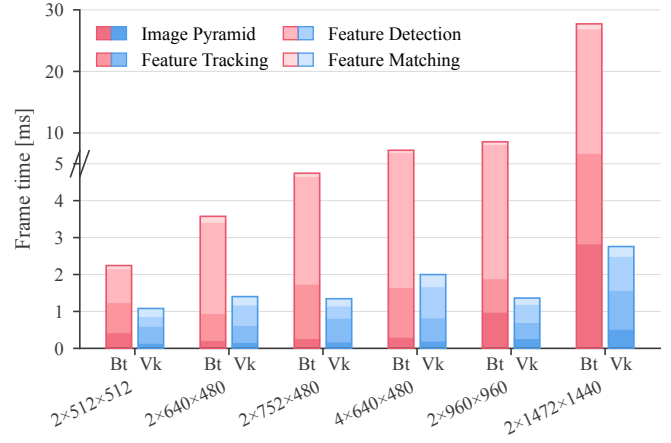}
    \caption{Per-stage frontend time of Basalt (Bt) and VkVIO (Vk) on the evaluation sequences, ordered from fewer to more input pixels. Stages include image pyramid, feature tracking, feature detection, and feature matching.}
    \label{fig:stages}
\end{figure}
Fig.~\ref{fig:stages} breaks the frontend into stages for the six evaluation sequences. Basalt's bars are host-side intervals between stage flags, whereas VkVIO's come from GPU-side timestamps that exclude command submission, the fence wait, and host-side overhead. Feature detection dominates Basalt's frontend (41\% of its stage time on TR2 and 62-73\% elsewhere) and rises fastest, from 0.9 to 20.2 ms. On the GPU, detection averages below 1 ms on every sequence. Basalt runs each cell sequentially on the CPU, whereas VkVIO dispatches cells across many GPU workgroups in parallel.

\begin{figure}[h]
    \centering
    \includegraphics[width=\columnwidth]{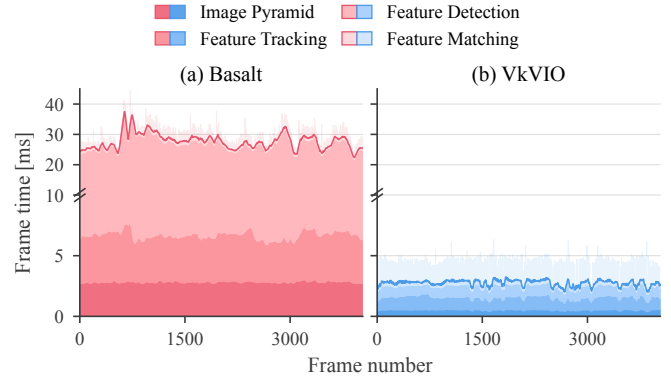}
    \caption{Per-frame frontend time of Basalt and VkVIO on HF151, illustrating the more stable timing of the GPU frontend. Stacked areas show the 50-frame rolling mean of each stage.}
    \label{fig:timing}
\end{figure}

Fig.~\ref{fig:timing} shows the per-frame frontend time over the whole HF151 sequence, Basalt in panel (a) from host stage flags and VkVIO in panel (b) from GPU-side timestamps. Per-frame times are not only faster on average but also far more stable: Basalt's frame time varies between roughly 20 and 45 ms, with feature detection causing about 95\% of this variation, while VkVIO stays within a band of about 5 ms. Consistent with this, the 99th-percentile end-to-end frame time in Table~\ref{tab:crossplatform} drops from 42.98 to 12.99 ms. Such a wide spread is costly for a real-time system because its latency budget has to cover the slowest frames rather than the average one.
\begin{table*}[t]
    \centering
    \scriptsize
    \caption{Cross-Platform Runtime Comparison}
    \label{tab:crossplatform}
    \setlength{\tabcolsep}{2pt}
\resizebox{\textwidth}{!}{%
    \begin{tabular}{llcccccccccccc}
        \toprule
                 &             & \multicolumn{2}{c}{\textbf{FPS}} & \multicolumn{3}{c}{\textbf{FE}} & \multicolumn{2}{c}{\textbf{BE}}     & \multicolumn{3}{c}{\textbf{E2E}}     & \multicolumn{2}{c}{\textbf{p99}}                                                                                                                                                    \\
        \cmidrule(lr){3-4}\cmidrule(lr){5-7}\cmidrule(lr){8-9}\cmidrule(lr){10-12}\cmidrule(lr){13-14}
        \textbf{Platf.} & \textbf{Seq.}        & Bt                      & Vk                     & Bt                         & Vk                          & Spd.                    & Bt                       & Vk                         & Bt                & Vk                          & Spd.        & Bt     & Vk              \\
        \midrule
        \multirow{6}{*}{\makecell{Apple                                                                                                                                                                                                                                                                                   \\ M3}}
                 & TR2         & 283                     & \textbf{351}           & $2.26\pm0.16$              & {\boldmath$1.63\pm0.17$}    & 1.4$\times$             & $1.27\pm0.44$            & {\boldmath$1.22\pm0.46$}   & $3.53\pm0.46$     & {\boldmath$2.85\pm0.51$}    & 1.2$\times$ & 5.26   & \textbf{4.59}   \\
                 & MOO02       & 209                     & \textbf{326}           & $3.61\pm0.37$              & {\boldmath$1.94\pm0.06$}    & 1.9$\times$             & $1.17\pm0.72$            & {\boldmath$1.13\pm0.72$}   & $4.77\pm0.71$     & {\boldmath$3.06\pm0.71$}    & 1.6$\times$ & 6.76   & \textbf{5.23}   \\
                 & EMH02       & 130                     & \textbf{204}           & $4.76\pm1.32$              & {\boldmath$1.93\pm0.23$}    & 2.5$\times$             & {\boldmath$2.95\pm1.60$} & $2.96\pm1.47$              & $7.71\pm1.67$     & {\boldmath$4.89\pm1.55$}    & 1.6$\times$ & 13.18  & \textbf{9.43}   \\
                 & MGO02       & 114                     & \textbf{237}           & $7.22\pm0.51$              & {\boldmath$2.79\pm0.67$}    & 2.6$\times$             & $1.54\pm0.85$            & {\boldmath$1.44\pm0.83$}   & $8.76\pm0.90$     & {\boldmath$4.23\pm1.08$}    & 2.1$\times$ & 10.91  & \textbf{6.95}   \\
                 & MIO02       & 101                     & \textbf{300}           & $8.58\pm0.47$              & {\boldmath$2.11\pm0.51$}    & 4.1$\times$             & $1.34\pm0.76$            & {\boldmath$1.23\pm0.72$}   & $9.92\pm0.78$     & {\boldmath$3.34\pm0.87$}    & 3.0$\times$ & 11.90  & \textbf{5.53}   \\
                 & HF151       & 32                      & \textbf{122}           & $27.78\pm2.96$             & {\boldmath$4.09\pm0.99$}    & 6.8$\times$             & {\boldmath$3.82\pm1.43$} & $4.11\pm1.54$              & $31.60\pm3.45$    & {\boldmath$8.20\pm1.84$}    & 3.9$\times$ & 42.98  & \textbf{12.99}  \\
        \midrule
        \multirow{6}{*}{\makecell{Radxa                                                                                                                                                                                                                                                                                   \\Cubie\\A7Z}}
                 & TR2         & 18                      & \textbf{21}            & {\boldmath$34.37\pm11.38$} & $34.62\pm2.47$              & 1.0$\times$             & $22.26\pm11.23$          & {\boldmath$12.67\pm4.60$}  & $56.63\pm17.98$   & {\boldmath$47.29\pm5.68$}   & 1.2$\times$ & 110.94 & \textbf{65.58}  \\
                 & MOO02       & 11                      & \textbf{17}            & $63.13\pm20.22$            & {\boldmath$50.52\pm3.14$}   & 1.2$\times$             & $24.81\pm20.15$          & {\boldmath$9.69\pm5.97$}   & $87.95\pm32.19$   & {\boldmath$60.22\pm6.20$}   & 1.5$\times$ & 189.07 & \textbf{77.90}  \\
                 & EMH02       & 7                       & \textbf{12}            & $71.17\pm30.44$            & {\boldmath$53.93\pm10.63$}  & 1.3$\times$             & $66.35\pm47.47$          & {\boldmath$26.13\pm13.86$} & $137.52\pm70.92$  & {\boldmath$80.06\pm22.64$}  & 1.7$\times$ & 364.30 & \textbf{142.74} \\
                 & MGO02       & 7                       & \textbf{8}             & $118.12\pm35.16$           & {\boldmath$103.68\pm5.46$}  & 1.1$\times$             & $33.93\pm25.21$          & {\boldmath$19.21\pm12.48$} & $152.06\pm46.74$  & {\boldmath$122.89\pm13.11$} & 1.2$\times$ & 278.04 & \textbf{165.99} \\
                 & MIO02       & 7                       & \textbf{9}             & $125.96\pm33.94$           & {\boldmath$100.04\pm7.23$}  & 1.3$\times$             & $24.78\pm17.58$          & {\boldmath$15.98\pm10.05$} & $150.74\pm39.93$  & {\boldmath$116.02\pm10.72$} & 1.3$\times$ & 248.92 & \textbf{146.37} \\
                 & HF151       & 2                       & \textbf{3}             & $406.45\pm72.81$           & {\boldmath$280.44\pm20.12$} & 1.4$\times$             & $91.63\pm59.04$          & {\boldmath$49.39\pm18.77$} & $498.08\pm106.76$ & {\boldmath$329.83\pm27.27$} & 1.5$\times$ & 760.47 & \textbf{397.93} \\
        \midrule
        \multirow{6}{*}{\makecell{Nvidia                                                                                                                                                                                                                                                                                  \\ RTX \\ 3070}}
                 & TR2         & 256                     & \textbf{550}           & $2.42\pm0.35$              & {\boldmath$0.50\pm0.05$}    & 4.8$\times$             & $1.49\pm0.45$            & {\boldmath$1.32\pm0.37$}   & $3.91\pm0.58$     & {\boldmath$1.82\pm0.38$}    & 2.2$\times$ & 5.67   & \textbf{3.21}   \\
                 & MOO02       & 179                     & \textbf{526}           & $4.07\pm0.64$              & {\boldmath$0.61\pm0.05$}    & 6.7$\times$             & $1.52\pm0.79$            & {\boldmath$1.29\pm0.67$}   & $5.59\pm0.91$     & {\boldmath$1.90\pm0.67$}    & 2.9$\times$ & 8.03   & \textbf{3.73}   \\
                 & EMH02       & 111                     & \textbf{286}           & $5.45\pm1.29$              & {\boldmath$0.58\pm0.07$}    & 9.5$\times$             & $3.54\pm1.76$            & {\boldmath$2.92\pm1.31$}   & $8.98\pm1.95$     & {\boldmath$3.49\pm1.35$}    & 2.6$\times$ & 15.42  & \textbf{7.19}   \\
                 & MGO02       & 103                     & \textbf{372}           & $7.77\pm0.91$              & {\boldmath$1.07\pm0.05$}    & 7.3$\times$             & $1.94\pm0.95$            & {\boldmath$1.62\pm0.77$}   & $9.71\pm1.32$     & {\boldmath$2.69\pm0.77$}    & 3.6$\times$ & 13.23  & \textbf{4.71}   \\
                 & MIO02       & 101                     & \textbf{394}           & $8.23\pm0.94$              & {\boldmath$1.05\pm0.07$}    & 7.8$\times$             & $1.62\pm0.77$            & {\boldmath$1.49\pm0.72$}   & $9.86\pm1.14$     & {\boldmath$2.54\pm0.71$}    & 3.9$\times$ & 12.78  & \textbf{4.32}   \\
                 & HF151       & 31                      & \textbf{156}           & $27.05\pm3.31$             & {\boldmath$2.40\pm0.11$}    & 11.3$\times$            & $4.73\pm1.48$            & {\boldmath$4.01\pm1.31$}   & $31.79\pm3.85$    & {\boldmath$6.42\pm1.33$}    & 5.0$\times$ & 42.90  & \textbf{10.19}  \\

        \bottomrule
    \end{tabular}%
}

\vspace{2pt}
\scriptsize\textit{
    \textbf{Bt}~Basalt\quad
    \textbf{Vk}~VkVIO\quad
    \textbf{Spd.}~VkVIO speedup\quad
    \textbf{FPS}~frames per second\quad
    \textbf{p99}~99th-percentile E2E frame time\\
    \textbf{FE/BE/E2E}~frontend/backend/end-to-end time per frame, mean\,$\pm$\,standard deviation\quad
    }
\end{table*}

Table~\ref{tab:crossplatform} repeats the comparison on three devices from different vendors. An Apple M3 laptop (Vulkan over Metal via MoltenVK), an Nvidia RTX 3070 desktop GPU, and the low-cost Radxa Cubie A7Z SBC. The same Vulkan pipeline runs on all three without vendor-specific code, only the Radxa, whose GPU lacks subgroup operations, swaps in our fallback shaders. On every device and every sequence, VkVIO lowers the end-to-end time, and its frontend time is lower or at parity. On the M3, the frontend speedup rises from 1.4$\times$ on TR2 to 6.8$\times$ on HF151 (E2E 1.2-3.9$\times$). On the RTX 3070, it ranges from 4.8$\times$ to 11.3$\times$ (E2E 2.2-5.0$\times$), with a frontend as short as 0.50 ms per frame and a per-frame standard deviation of 0.05-0.11 ms against 0.35-3.31 ms for Basalt. The Radxa gains 1.0-1.4$\times$ in the frontend and 1.2-1.7$\times$ end to end, partly because its backend runs 1.6-2.6$\times$ faster, as the frontend no longer competes for the CPU cores.
\begin{figure}[h]
    \centering
    \includegraphics[width=\columnwidth]{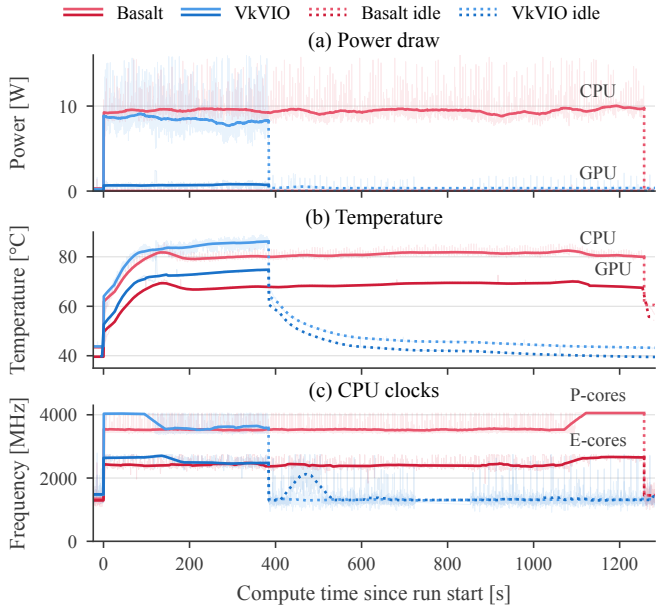}
    \caption{Long-term operation on the MIPB08 endurance sequence on the M3. We show power, temperature, and clock frequency for the CPU (Basalt) and GPU (VkVIO) frontends. Raw samples are drawn shaded under a centered 50~s rolling mean.}
    \label{fig:longterm}
\end{figure}
Fig.~\ref{fig:longterm} shows long-term operation on MIPB08, a 36.5-minute session, on the Apple M3. VkVIO completes the session in 384~s against 1257~s for Basalt, at the same accuracy (ATE 0.603~m vs.\ 0.597~m) and at a lower power of 9.9~W against 10.3~W. It therefore needs 3.4$\times$ less energy for the session, 32~mJ per frame against 110~mJ. The GPU frontend heats the SoC slightly more than the CPU baseline while running. VkVIO holds its P-cores at their 4.04~GHz peak for the first 116~s, while Basalt's fluctuate around 3.5~GHz for most of the run. Only once the temperature peaks at 84~$^\circ$C do VkVIO's P-cores drop to about 3.6~GHz, without affecting its per-frame time.

\begin{table*}[t]
    \centering
    \scriptsize
    \caption{Comparison with CUDA Frontends}
    \label{tab:cuda}
    \resizebox{\textwidth}{!}{%
  \setlength{\tabcolsep}{2.5pt}
  \begin{tabular}{lcccccccccccccccc}
    \toprule
        & \multicolumn{4}{c}{\textbf{FPS}} 
        & \multicolumn{4}{c}{\textbf{E2E}} 
        & \multicolumn{4}{c}{\textbf{ATE}} 
        & \multicolumn{4}{c}{\textbf{RTE}} \\
                
        \cmidrule(lr){2-5}
        \cmidrule(lr){6-9}
        \cmidrule(lr){10-13}
        \cmidrule(lr){14-17}
        
    \textbf{Seq.}
      & Vk & Bt & JS & FT                      
      & Vk & Bt & JS & FT           
      & Vk & Bt & JS & FT             
      & Vk & Bt & JS & FT    
    \\
    
    \midrule
    
    TR2
    & \textbf{535} & 287 & 144 & 177
    & {\boldmath$1.87\pm0.37$} & $3.49\pm0.40$ & $6.93\pm2.76$ & $5.66\pm0.89$
    & 0.053 & 0.059 & 0.146 & \textbf{0.026}
    & 0.007 & \textbf{0.007} & 0.117 & 0.011
    \\

    MOO02
    & \textbf{495} & 193 & 159 & 158
    & {\boldmath$2.02\pm0.70$} & $5.18\pm0.75$ & $6.30\pm2.79$ & $6.30\pm1.05$
    & 0.235 & \textbf{0.234} & 0.392 & 0.348
    & \textbf{0.015} & 0.015 & 0.214 & 0.042
    \\

    EMH02
    & \textbf{269} & 128 & 136 & 172
    & {\boldmath$3.72\pm1.40$} & $7.82\pm1.47$ & $7.33\pm4.43$ & $5.82\pm1.47$
    & \textbf{0.049} & 0.055 & 0.051 & 0.145
    & \textbf{0.004} & 0.004 & 0.009 & 0.050
    \\

    MGO02\textsuperscript{$\dagger$}
    & \textbf{334} & 112 & 197 & 206
    & {\boldmath$2.99\pm0.83$} & $8.90\pm0.91$ & $5.07\pm2.08$ & $4.86\pm1.23$
    & 0.503 & \textbf{0.497} & 0.654 & 1.922
    & \textbf{0.028} & 0.028 & 0.246 & 0.756
    \\

    MIO02
    & \textbf{379} & 110 & 98 & 150
    & {\boldmath$2.64\pm0.70$} & $9.06\pm1.00$ & $10.20\pm2.38$ & $6.66\pm1.20$
    & 1.134 & 1.133 & \textbf{0.197} & 5.410
    & \textbf{0.056} & 0.058 & 0.100 & 0.191
    \\

    HF151
    & \textbf{145} & 34 & \cxmark & \cxmark
    & {\boldmath$6.90\pm1.45$} & $29.37\pm3.38$ & \cxmark & \cxmark
    & 0.904 & \textbf{0.901} & \cxmark & \cxmark
    & \textbf{0.374} & 0.375 & \cxmark & \cxmark
    \\
    \bottomrule
  \end{tabular}%
}

\vspace{2pt}

\scriptsize\textit{
  \textbf{Vk}~VkVIO \quad
  \textbf{Bt}~Basalt \quad
  \textbf{JS}~Jetson-SLAM \quad
  \textbf{FT}~FastTrack \quad
  \textbf{FPS}~frames per second \quad
  \textbf{E2E}~end-to-end time per frame, mean\,$\pm$\,standard deviation in ms \quad
  \textbf{ATE/RTE}~absolute/relative trajectory error. $SE(3)$ aligned/$\Delta=6$ frames\quad
  \textsuperscript{$\dagger$}\,Jetson-SLAM and FastTrack process MGO02 with only 2 of the 4 cameras \quad
  \xmark~run failed\quad
}
\end{table*}
\subsection{Improvements over CUDA-Accelerated Systems}
Having shown that VkVIO outperforms the CPU baseline on GPUs of three vendors, we now ask whether a vendor-agnostic API costs performance on Nvidia hardware itself. Table~\ref{tab:cuda} compares VkVIO, Basalt, and the CUDA-only frontends Jetson-SLAM~\cite{kumarHighSpeedStereoVisual2024} and FastTrack~\cite{khabiriFastTrackGPUAcceleratedTracking2025a} on one desktop with an Intel Core i5-13600 CPU and an Nvidia RTX 3070 GPU. Jetson-SLAM builds on ORB-SLAM2 and tracks two cameras without inertial data, while FastTrack builds on ORB-SLAM3 and leverages the IMU too. Neither supports the four cameras of MGO02, and neither completes HF151, whose two cameras face opposite directions. Both run mapping and loop closing in separate threads, so their end-to-end times cover the per-frame tracking path only, and we evaluate the pose each reports online rather than its post-optimization trajectory.
\begin{figure}[b]
    \centering
    \includegraphics[width=\columnwidth]{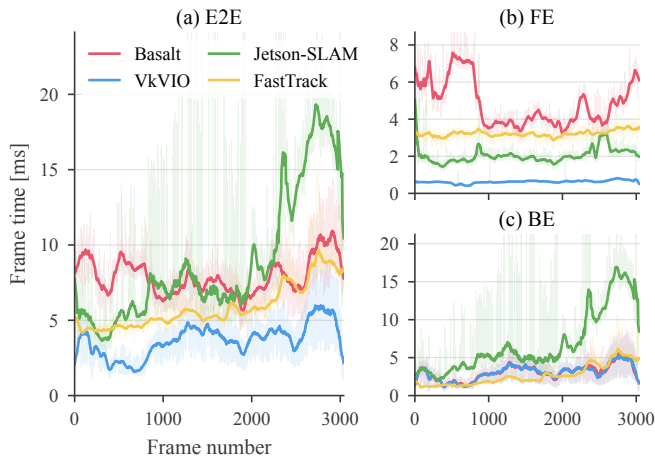}
    \caption{End-to-end, frontend, and backend times of VkVIO against Basalt and the CUDA-based FastTrack and Jetson-SLAM on EMH02. Raw per-frame times are drawn shaded under a centered 50-frame rolling mean.}
    \label{fig:cudatimes}
\end{figure}

VkVIO reaches the highest frame rate of the four systems on every sequence. Its end-to-end time is 1.6-3.1$\times$ faster than FastTrack's, 1.7-3.9$\times$ lower than Jetson-SLAM's, and 1.9-4.3$\times$ than Basalt's, with the smallest frame-to-frame spread of the four (0.37-1.45 ms against 0.89-1.47 ms for FastTrack and 2.1-4.4 ms for Jetson-SLAM). Fig.~\ref{fig:cudatimes} shows where this margin comes from on EMH02. VkVIO's frontend averages 0.63 ms per frame against 2.04 ms for Jetson-SLAM, 3.22 ms for FastTrack, and 4.83 ms for Basalt. VkVIO and Basalt share the same sliding-window optimization at about 3.1 ms, FastTrack's tracking thread takes 2.6 ms, and Jetson-SLAM's grows with its map to over 14 ms in the last 500 frames.

Accuracy remains that of Basalt, within 6 mm of ATE and 2 mm of RTE on all six sequences. Against the CUDA baselines, VkVIO has the lower ATE on three of the five sequences they complete, and the lower RTE on all five. FastTrack has the lower ATE on TR2 and Jetson-SLAM on MIO02, where the shared Basalt backend drifts to over 1 m. To our knowledge, GLidE-SLAM~\cite{desousaGLidESLAMGLAcceleratedIndirectDirect2026a} is the only other vendor-agnostic GPU-accelerated tracking frontend, we exclude it from our evaluation since it is monocular only and no public implementation was available at the time of writing.

\subsection{Deployment on a Low-Cost and Low-Power Platform}
The Radxa Cubie A7Z is a device with low cost and power. It is a single-board computer with a 5-10 W power envelope that costs about 20$\times$ less than the closest CUDA-capable device Nvidia still maintains. We use the bare board, without a heatsink or fan, which makes heat the first constraint.

\begin{figure}[h]
    \centering
    \includegraphics[width=\columnwidth]{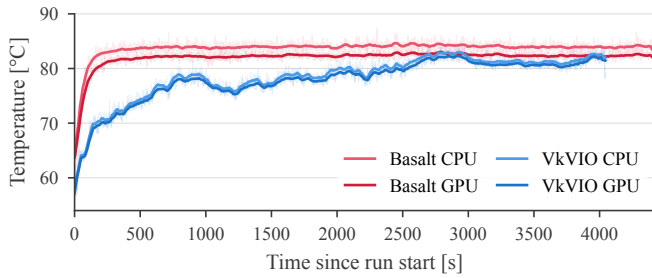}
    \caption{CPU and GPU temperatures for Basalt and VkVIO on the Radxa
    Cubie A7Z over MOO12, run with no heatsink and no fan. Raw samples are drawn
    shaded under a centered 50~s rolling mean. VkVIO's traces
    stop earlier because it finishes the session sooner.}
    \label{fig:radxatemps}
\end{figure}

Fig.~\ref{fig:radxatemps} shows its CPU and GPU temperatures when running a long 40-minute sequence (MOO12) as fast as possible. Basalt pushes the CPU past 80~$^\circ$C within 80\,s and holds it at 84~$^\circ$C for the rest of the run while VkVIO needs 13 minutes to reach 80~$^\circ$C, stays cooler from then on (by 5~$^\circ$C on average and 2~$^\circ$C at the end), and finishes earlier.

\begin{table}[h]
    \centering
    \footnotesize
    \caption{VkVIO-lite on the Radxa}
    \label{tab:radxalite}
    \resizebox{\columnwidth}{!}{%
  \setlength{\tabcolsep}{2.5pt}
  \begin{tabular}{llccccccc}
    \toprule
    \textbf{Seq.}          & \textbf{Sys.} & \textbf{FPS} & \textbf{FE}              & \textbf{BE}              & \textbf{E2E}              & \textbf{p99}   & \textbf{ATE}   & \textbf{RTE}   \\
    \midrule
    \multirow{3}{*}{TR2}   & Bt            & 18           & $34.2\pm11.3$            & $21.5\pm11.0$            & $55.6\pm17.8$             & 109.6          & \textbf{0.068} & \textbf{0.008} \\
                           & Vk            & 22           & $34.5\pm2.5$             & $12.1\pm4.5$             & $46.6\pm5.6$              & 64.9           & 0.068          & 0.008          \\
                           & Vk$_L$        & \textbf{35}  & {\boldmath$18.8\pm1.1$}  & {\boldmath$9.8\pm4.1$}   & {\boldmath$28.6\pm4.6$}   & \textbf{42.5}  & 0.082          & 0.008          \\
    \midrule
    \multirow{3}{*}{MOO02} & Bt            & 12           & $62.6\pm20.1$            & $23.8\pm19.9$            & $86.4\pm31.9$             & 184.6          & 0.243          & 0.015          \\
                           & Vk            & 17           & $50.2\pm3.1$             & {\boldmath$9.2\pm5.8$}   & $59.5\pm6.1$              & 76.9           & 0.240          & 0.015          \\
                           & Vk$_L$        & \textbf{32}  & {\boldmath$20.8\pm1.7$}  & $10.0\pm4.6$             & {\boldmath$30.8\pm4.8$}   & \textbf{42.7}  & \textbf{0.235} & \textbf{0.015} \\
    \midrule
    \multirow{3}{*}{EMH02} & Bt            & 7            & $71.1\pm30.4$            & $65.2\pm46.8$            & $136.2\pm70.2$            & 361.4          & \textbf{0.048} & \textbf{0.004} \\
                           & Vk            & 13           & $53.8\pm10.6$            & $25.5\pm13.6$            & $79.4\pm22.4$             & 141.4          & 0.055          & 0.004          \\
                           & Vk$_L$        & \textbf{24}  & {\boldmath$25.7\pm3.6$}  & {\boldmath$16.3\pm7.6$}  & {\boldmath$41.9\pm10.2$}  & \textbf{79.4}  & 0.058          & 0.005          \\
    \midrule
    \multirow{3}{*}{MGO02} & Bt            & 7            & $117.6\pm35.1$           & $32.4\pm24.8$            & $150.1\pm46.4$            & 275.4          & 0.568          & 0.028          \\
                           & Vk            & 8            & $103.3\pm5.5$            & $18.1\pm12.1$            & $121.4\pm12.8$            & 163.0          & 0.570          & 0.028          \\
                           & Vk$_L$        & \textbf{18}  & {\boldmath$40.6\pm2.9$}  & {\boldmath$16.0\pm9.6$}  & {\boldmath$56.6\pm10.1$}  & \textbf{90.5}  & \textbf{0.361} & \textbf{0.026} \\
    \midrule
    \multirow{3}{*}{MIO02} & Bt            & 7            & $125.7\pm33.9$           & $23.9\pm17.3$            & $149.5\pm39.8$            & 247.6          & 1.165          & \textbf{0.063} \\
                           & Vk            & 9            & $99.8\pm7.2$             & $15.3\pm9.9$             & $115.1\pm10.5$            & 144.9          & \textbf{1.163} & 0.064          \\
                           & Vk$_L$        & \textbf{15}  & {\boldmath$53.8\pm1.6$}  & {\boldmath$14.5\pm7.7$}  & {\boldmath$68.2\pm8.2$}   & \textbf{92.9}  & 1.661          & 0.085          \\
    \midrule
    \multirow{3}{*}{HF151} & Bt     & 2           & $406.1\pm72.7$            & $88.4\pm57.8$             & $494.6\pm106.0$            & 753.8          & 1.269          & \textbf{0.374} \\
                           & Vk            & 3            & $279.4\pm20.1$           & $45.9\pm17.5$            & $325.3\pm26.4$            & 391.2          & \textbf{0.821} & 0.374          \\
                           & Vk$_L$        & \textbf{6}   & {\boldmath$129.3\pm4.3$} & {\boldmath$31.6\pm10.8$} & {\boldmath$160.9\pm11.9$} & \textbf{195.6} & 1.618          & 0.377          \\
    \bottomrule
  \end{tabular}%
}

\vspace{2pt}

\scriptsize\textit{
  \textbf{ATE}~$SE(3)$ aligned\quad
  \textbf{RTE}~$\Delta=6$ frames\quad
  \textbf{p99}~99th-percentile E2E frame time\quad
  \textbf{FE/BE/E2E}~front/back/end-to-end mean times per frame\quad
  \textbf{$\pm$}~standard deviation\quad
  \textbf{Bt}~Basalt\quad
  \textbf{Vk}~VkVIO\quad
  \textbf{Vk$_L$}~VkVIO-lite\quad
  \textbf{FPS}~frames per second\quad
  }
\end{table}

Throughput is the second constraint. VkVIO's standard configuration lowers Basalt's frame times on every sequence of Table~\ref{tab:radxalite}, as in Table~\ref{tab:crossplatform}, but only TR2 reaches its camera rate. For this board, we therefore define VkVIO-lite. It raises the minimum FAST threshold from 5 to 20, increases the cell size from 50 to 82, and reduces the Gauss-Newton iterations per level from 5 to 3. For the backend, we decrease the optimization window from 7 keyframes to 6. The frontend speeds up a further 1.8-2.5$\times$, and TR2, MOO02, and EMH02 now run above their camera rates on average, with 99th-percentile frame times 2.6-4.6$\times$ shorter than Basalt's. MGO02, MIO02, and HF151 still fall short. ATE stays within 15 mm of the standard configuration on the three real-time sequences and improves on MGO02, but grows by 0.5 m on MIO02 and 0.8 m on HF151. RTE increases noticeably only on MIO02 by 21\,mm.

\section{CONCLUSIONS}

We presented VkVIO, to the best of our knowledge the first cross-platform GPU-accelerated stereo-inertial VIO frontend, implemented in Vulkan on top of Basalt. Across Apple, Nvidia, and Radxa GPUs, VkVIO delivers consistent frontend and end-to-end speedups over Basalt's CPU implementation while preserving tracking accuracy (Tab.~\ref{tab:accuracy} and~\ref{tab:crossplatform}). Beyond raw speed, VkVIO's frame times are far more stable (Fig.~\ref{fig:timing}, Tab.~\ref{tab:crossplatform}), and the GPU frontend presents better power efficiency and thermals (Figs.~\ref{fig:longterm} and~\ref{fig:radxatemps}) over long operation. VkVIO beats existing CUDA-only accelerated frontends (Jetson-SLAM, FastTrack) on the same Nvidia hardware, despite going through a vendor-agnostic API rather than a vendor-specific one (Tab.~\ref{tab:cuda}, Fig.~\ref{fig:cudatimes}).

We showed that VkVIO makes real-time VIO viable (Tab.~\ref{tab:radxalite}) on an extremely inexpensive, low-power SBC, $\sim$20$\times$ cheaper than the closest CUDA-capable device Nvidia maintains, a platform none of the CUDA baselines can even run on. By removing the CUDA lock-in, VkVIO gives practitioners the flexibility to prototype on one GPU vendor and deploy on another without rewriting the frontend, and to pick the best available hardware for each application's power, size, and thermal budget.

We hope this work encourages wider adoption of Vulkan as a serious, vendor-neutral compute target for real-time robotics and XR perception pipelines, beyond its traditional role in graphics.

\bibliographystyle{IEEEtran}
\balance
\bibliography{IEEEabrv,main}

\immediate\closein\imgstream

\end{document}